%% file: journal.tex
\documentclass[journal]{IEEEtran}
\usepackage{standalone} % needs to be on top

\usepackage{amsmath,amsfonts,amsthm}
\usepackage{algorithmic}
\usepackage{algorithm}
\usepackage{array}
\usepackage[caption=false,font=normalsize,labelfont=sf,textfont=sf]{subfig}
\usepackage{textcomp}
\usepackage{stfloats}
\usepackage{url}
\usepackage{verbatim}
\usepackage{graphicx}
\usepackage{cite}
\usepackage{enumitem}

\usepackage{tikz}
\usetikzlibrary{positioning}
\usetikzlibrary{calc}

\usepackage{bm}
\usepackage{bbm}
\usepackage{booktabs}
\usepackage{multirow}
\usepackage{xcolor}
\usepackage{hyperref}
\usepackage{style/macros}

\newtheorem{definition}{Definition}
\newtheorem{lemma}{Lemma}
\newtheorem{theorem}{Theorem}
\newtheorem{assumption}{Assumption}

\newcommand{\figWidth}{0.85}
\newcommand{\inW}{A}
\newcommand{\outW}{B}
\newcommand{\Linf}{\calL_\infty}
\newcommand{\Lfin}{\calL_\sqcap}
\newcommand{\norm}[1]{\|#1\|}
\newcommand{\avg}[1]{ \lim_{T \to \infty} \frac{1}{T^d} \int_{\calC_T^d} #1 d\bft}

\title{Transferability of Convolutional Neural Networks in Stationary Learning Tasks}
\author{
Damian Owerko, 
Charilaos I. Kanatsoulis,
Jennifer Bondarchuk,
Donald J. Bucci Jr, and
Alejandro Ribeiro
\thanks{Most experimental results in this paper have been previously presented in \cite{Owerko23-MultiTargetTracking} for multi-target tracking and \cite{Owerko23-SolvingLargescale} for mobile infrastructure on demand along with a summary of the theoretical result.}
\thanks{D. Owerko, C. I. Kanatsoulis, and A. Riberio are with the Department of Electrical and Systems Engineering, University of Pennsylvania, Philadelphia, USA.}
\thanks{J Bondarchuk and D. J. Bucci Jr are with the Advanced Technology Labs, Lockheed Martin, Cherry Hill, USA.}
}

\begin{document}
\bstctlcite{IEEEexample:BSTcontrol}

\maketitle

\begin{abstract}
    Recent advances in hardware and big data acquisition have accelerated the development of deep learning techniques. For an extended period of time, increasing the model complexity has led to performance improvements for various tasks. However, this trend is becoming unsustainable and there is a need for alternative, computationally lighter methods. In this paper, we introduce a novel framework for efficient training of convolutional neural networks (CNNs) for large-scale spatial problems. To accomplish this we investigate the properties of CNNs for tasks where the underlying signals are stationary. We show that a CNN trained on small windows of such signals achieves a nearly performance on much larger windows without retraining. This claim is supported by our theoretical analysis, which provides a bound on the performance degradation. Additionally, we conduct thorough experimental analysis on two tasks: multi-target tracking and mobile infrastructure on demand. Our results show that the CNN is able to tackle problems with many hundreds of agents after being trained with fewer than ten. Thus, CNN architectures provide solutions to these problems at previously computationally intractable scales.
\end{abstract}


\input{journal/introduction}
\input{journal/stationary.tex}
\input{journal/representing}

\input{journal/architecture.tex}
\input{journal/mtt.tex}
\input{journal/mid.tex}
\input{journal/conclusions.tex}
\appendices
\input{journal/appendix.tex}
\bibliography{IEEEabrv,bib/settings,bib/journal,bib/lmco}

\end{document}

%% file: journal/introduction.tex
\section{Introduction}
How much longer can we sustain the growing dataset size and model complexity? Over the past decade, we have seen rapid advancements in deep learning, which produced state-of-the-art results in a wide range of applications \cite{Shinde18-ReviewMachine, Gu18-RecentAdvances, Lin21-SurveyTransformers}. In large part, these successes are due to increasingly powerful hardware \cite{Shinde18-ReviewMachine, Gu18-RecentAdvances} that enabled processing larger datasets \cite{Najafabadi15-DeepLearning} and training deep learning models with more parameters. 

Recently, this trend has been taken to new extremes with the advent of large language models \cite{Brown20-LanguageModelsAre, Rae22-ScalingLanguageModels, Smith22-UsingDeepSpeed, Touvron23-LLaMAOpen}. For example, GPT-3 was trained on a dataset with approximately 374 billion words and has 175 billion parameters \cite{Brown20-LanguageModelsAre}. However, this strategy is not sustainable due to diminishing returns and increasing cost of computation and data acquisition \cite{Thompson22-Computational, Patterson21-CarbonEmissions}. Moreover, in applications with a limited amount of data, training such large models is impractical. A natural question that arises, is whether we can build computationally efficient models, that maintain the favorable properties of deep learning. In this paper, we give an affirmative answer and develop an analytical and computational framework that uses fully convolutional models and transfer learning.

Convolutional neural networks (CNNs) are one of the most popular deep learning architectures \cite{Gu18-RecentAdvances}, especially for image classification \cite{Rawat17-DeepConvolutional}. Though initially used for image processing, they have proven useful for a wide variety of other signals such as text, audio, weather, ECG data, traffic data, and many others \cite{Gu18-RecentAdvances, Li22-SurveyConvolutional, Alipour19-RobustPixelLevel}. Despite the recent increased interest in transformer-based models  \cite{Lin21-SurveyTransformers}, convolutions continue to play an important role under the hood of novel architectures. For example, the autoencoder in latent diffusion models \cite{Rombach22-HighResolution} is fully convolutional \cite{Esser21-TamingTransformers}.

Transfer learning, on the other hand, is a set of methodologies that exploit knowledge from one domain to another \cite{Zhuang21-ComprehensiveSurvey, Tan18-SurveyDeepTransfer}. The works in \cite{Ribani19-SurveyTransfer, Zhu22-TransferLearningDeep} taxonomize a wide range of transfer learning approaches. One common transfer learning approach is \emph{parameter transfer}, where a model is pre-trained to perform one task and then some or all of its parameters are reused for another task. This process allows a model to grasp intricate patterns and features present in one domain and apply that knowledge to a different, but related domain. For example, the work in \cite{Shin16-DeepConvolutional} employs parameter transfer to improve CNN performance on medical images by pre-training on ImageNet. Yet, the true power of transfer learning transcends mere performance improvements. It offers an opportunity to construct highly efficient and lightweight solutions, which is crucial in large-scale systems with high computational needs or constrained resources. In the context of graph neural networks, transfer learning has been explored, both theoretically and experimentally, for large-scale graphs \cite{Ruiz20-GraphonNeural}. Training on small graphs and transferring to larger graphs is a promising approach \cite{Gama21-GraphNeuralNetworks,Tzes23-GraphNeuralNetworks}. However, this flavor of transfer learning has been underexploited for CNNs, with a few exceptions; \cite{Madala23-CNNsAvoidCurse} shows that training on CIFAR can be decomposed into training on smaller patches of images.

In this paper, we fill this gap and develop a novel theoretical and computational framework to tackle large-scale spatial problems with transfer learning and CNNs. 
In particular, we train a CNN model on a small version of a spatial problem and transfer it to an arbitrarily large problem, without re-training. 
To justify this approach we leverage the shift-equivariance property of CNNs and analyze their performance at arbitrarily large scales. Note that, many machine learning tasks are shift-equivariant in some sense. For instance, in image segmentation, if the input image is translated the output will move accordingly. Other systems exhibit this phenomennon, including financial \cite{Rinn15-Dynamics}, weather \cite{Michelangeli95-Weather}, and multi-agent systems \cite{Owerko23-MultiTargetTracking}.

Our analysis studies CNNs when the input-output pairs are stationary processes over space or time. This is motivated by the fact that, in stochastic process theory, shift-invariance is equivalent to stationarity \cite{Kallenberg21-Foundations} and shift-equivariance is equivalent to joint stationarity of the input and output signals. We demonstrate that a CNN can be efficiently trained for tasks with jointly stationary inputs and outputs. To accomplish this, we prove a bound on the generalization error of a fully convolutional model when trained on small windows of the input and output signals. Our theory indicates that such a model can be deployed on arbitrarily large windows on the signals, with minimal performance degradation.

% Similarly, in \cite{Ruiz21-GraphNeuralNetworks} the authors show theoretically that graph neural networks (GNNs) are shift-equivariant and stable to pertubations of the graph, which allows weights can be resused to perform the same task on different graphs. This is demonstrated experimentally in \cite{Gama21-GraphNeuralNetworks,tzes2023graph} where the authors train a GNN training on small graphs and transfer to larger ones.

% Rephrase for clarity: This flavor of transfer learning has been underexploited for CNNs, with a few exceptions e.g., the authors of \cite{Madala23-CNNsAvoidCurse} show that shift-equivariance of CNNs has a regularizing effect on image classification performance. They show that training on CIFAR can be decomposed into training on smaller patches of images.

To support this analysis we conducted numerical experiments on two tasks: multi-target tracking \cite{Vo15-MultitargetTracking} and mobile infrastructure on demand \cite{Mox20-MobileWirelessNetwork, Mox22-Learning}. In both problems, the inputs and outputs are finite sets, which we reinterpret as image-to-image tasks. We describe this method in detail to show how to apply our framework to a broad class of spatial problems. Overall, our contributions can be summarized as follows.

\begin{enumerate}[label=\textbf{(C\arabic*)}]
    \item Present a transfer learning framework for efficient training of CNNs for large-scale spatial problems.
    \item Prove a bound for the generalization error of a CNN trained under this framework.
    \item Describe a methodology to apply CNNs to non-image problems.
    \item Demonstrate the effectiveness of the framework in multi-target tracking and mobile infrastructure on demand. 
\end{enumerate}

%% file: journal/stationary.tex
% !TeX root = journal.tex

\section{Learning to process stationary signals}\label{sec:process_stationary}
The following is a common machine learning (ML) problem that becomes challenging when signals have infinite support. Given an input-output signal pair \( (X,Y) \), the task is to find a model \(\bm\Phi(\cdot)\) that minimizes the mean squared error (MSE) between \( \bm\Phi(X) \) and \( Y \). Let \(X\) and \(Y\) be random functions of a vector parameter \(\bft \in \reals^d\), which can represent time, space, or an abstract \(d\)-dimensional space. Mathematically, this is a minimization problem with loss
\begin{equation}\label{eq:mse_canonical}
    \calL_1(\bm\Phi) = \E{ \int_{\reals^d} |\bm\Phi(X)(\bft) - Y(\bft)|^2 d\bft },
\end{equation}
where \(\bm\Phi(X)(\bft)\) is the output of the model at \(\bft\).
When the signals have infinite support, the integral may diverge. To accommodate this, the MSE can be redefined as,
\begin{equation}\label{eq:mse_general}
    \calL_2(\bm\Phi) = \E{ \avg{|\bm\Phi(X)(\bft) - Y(\bft)|^2}},
\end{equation}
where \( C^d_T \) is an open set of the points in \( \reals^d \) within a \(T\)-wide hypercube, centered around the origin. The primary difference between \eqref{eq:mse_canonical} and \eqref{eq:mse_general} is the normalization by the hypercube volume \(T^d\). This extends the MSE to infinite signals where \eqref{eq:mse_canonical} diverges. Practically, large values of \(T\) represent situations where signals are too wide to be computationally tractable.

This paper explores how to solve the problem of minimizing \eqref{eq:mse_general}. Numerically evaluating and minimizing \eqref{eq:mse_general} is intractable in general. To overcome this limitation, Section \ref{sec:window} shows that if \(\bm\Phi(\cdot)\) is a CNN trained on small windows of the signals, then \eqref{eq:mse_general} is bounded by the training loss of the small window plus a small quantity. To accomplish this \(X,~Y\) are modeled as jointly stationary random signals and the output of the CNN \(\bm\Phi(X)\) is analyzed. Before this analysis is presented in \ref{sec:window}, the following sections define stochastic processes, stationarity, and CNNs.

\subsection{Jointly Stationary Stochastic Processes}\label{subsec:stationary_process}

We focus on continuous stationary stochastic processes with a multidimensional index. In this context, a stochastic process \cite{Gray09-ProbabilityRandom, Kallenberg21-Foundations, Shalizi07-AlmostNoneTheory} is a family of random variables \( \{X(\bft)\}_{\bft \in \reals^d} \) indexed by a parameter \(\bft \in \reals^d\).
% A stochastic process \cite{Gray09-ProbabilityRandom, Kallenberg21-Foundations, RohillaShalizi07-AlmostNone} is a family of random variables \(\{X(t): \Omega \to S\}_{t \in \calI}\) parametrized by an index set \(\calI\). Each element in this family is a map from a sample space \(\Omega\) onto a measurable space \(S\). Equivalently, we can describe the process as a map \(X(t,\omega): \calI \times \Omega \to S\) from the cartesian product of the index set and the sample space onto the measurable space. In this context, a random signal is the corresponding random function, \(X := X(\cdot, \omega)\). These representations are equivalent and we will use them interchangeably.
One of the key assumptions in the analysis is joint stationary between signals \(X\) and \(Y\). 
% We assume that \(X\) and \(Y\) are two such stochastic processes, \(X: \reals^d \times \Omega \to \reals\) and \(Y: \reals^d \times \Omega \to \reals\). Our key assumption is that the two processes are jointly stationary. 
Two continuous random signals are jointly stationary if all their finite joint distributions are shift-invariant.
\begin{definition}\label{def:jointly_stationary}
    Consider two real continuous stochastic processes \(\{X(\bft)\}_{\bft\in\reals^d}\) and \(\{Y(\bft)\}_{\bft\in\reals^d}\). The two processes are jointly stationary if and only if they satisfy \eqref{eq:jointly_stationary} for any shift \(\bm\tau \in \reals^d\), non-negative \(n, m \in \mathbb{N}_0\), indices \(\bft_i, \bfs_i \in \reals^d\), and Borel sets of the real line \(A_i, B_i\).
    \begin{equation}
        \begin{split}\label{eq:jointly_stationary}
            &P(X(\bft_1) \in A_1, ..., X(\bft_n) \in A_n, ...,  Y(\bfs_m) \in B_m)\\
            &=P(X(\bft_1+\bm\tau), ..., X(\bft_n+\bm\tau) \in A_n, ..., Y(\bfs_m+\bm\tau) \in B_m)
        \end{split}
    \end{equation}
\end{definition}

Modeling \(X\) and \(Y\) as jointly stationary stochastic processes in \(d\)-dimensions has practical applications in many domains.  Numerous signals can be represented by stationary or quasi-stationary processes. Examples include signals in financial \cite{Rinn15-Dynamics}, weather \cite{Michelangeli95-Weather}, sensor network \cite{Xie11-AnomalyDetection} and multi-agent systems \cite{Owerko23-MultiTargetTracking}. Therefore, understanding the performance of ML models for stationary signal processing has widespread importance. CNNs are of particular interest because they admit translational symmetries similar to jointly stationary signals. The following analysis attempts to demystify their performance and broaden our knowledge of how to train CNNs efficiently.

\subsection{Convolutional Neural Networks}\label{subsec:cnn}

CNNs are versatile architectures, heavily used to minimize \eqref{eq:mse_canonical} -- typically to perform image processing. A CNN is a cascade of non-linear functions, called layers. The output of the \( l^\text{th} \) is defined recursively:
\begin{equation}\label{eq:cnn}
    x_l(\bft) = \sigma \left( \int_{\reals^d} h_l(\bfs) x_{l-1}(\bft - \bfs) d\bfs \right).
\end{equation}
At each layer, the output \(x_l\) is obtained by convolving the previous output $x_{l-1}$ by a function \(h_{l}\) and applying a pointwise nonlinearity \(\sigma\). As the name suggests, convolution operations are the cornerstone of the architecture. They are the key to exploiting spatial symmetries since they are shift-equivariant. 
The CNN has \(L\) layers and is parametrized by a set of functions \(\calH = \{ h_1,...,h_L\}\), also known as parameters. Let
\begin{equation}
    \bm\Phi(X; \calH)(t) \defeq x_L(t)
\end{equation}
represent a CNN with output \(x_L\) and input \(x_0 \defeq X\). Then the loss function in \eqref{eq:mse_general} can be rewritten as a function of these parameters.
\begin{equation}\label{eq:mse_inf}
    \Linf(\calH) = \E{ \avg{ |\bm\Phi(X; \calH)(\bft) - Y(\bft)|^2 } }
\end{equation}

In general, evaluating \eqref{eq:mse_inf} is as challenging as evaluating \eqref{eq:mse_general}, but the following two assumptions make this possible. First, the input and output signals have a stationary property. Second, the model \( \bm\Phi \) is a CNN. The next section presents our main theoretical result, which discusses how the CNN can be efficiently trained on narrow signal windows, but evaluated on arbitrarily wide signals without sacrificing performance.

\subsection{Training on a window}\label{sec:window}

Instead of minimizing \eqref{eq:mse_inf} directly, which is computationally prohibitive, we propose to train the CNN on narrow windows of the signals \(X,~Y\). In particular, we can rewrite \eqref{eq:mse_inf} to consider finite windows over the signals:
\begin{equation}\label{eq:mse_window}
    \Lfin(\calH) = \frac{1}{\outW^d}
    \E{ \int_{\calC_\outW^d} | \bm\Phi(\sqcap_\inW X; \calH)(\bft) - Y(\bft) |^2  dt },
\end{equation}
where $\sqcap_s(\bft) \defeq \mathbbm{1}(\bft \in \calC_s^d)$ is an indicator function that is equal to one whenever \(\bft\) is within a \(s\)-wide hypercube. Minimizing \eqref{eq:mse_window} is equivalent to learning a model between \( \sqcap_\inW X \) and  \( \sqcap_\outW Y \). These are windowed versions of the original signals, with window widths \(\inW\) and \(\outW\), respectively. We will only consider \(\inW \ge \outW\) so that the input is wider than the output of the CNN. Most practical CNN architectures either shrink their input width or keep it constant because padding at each convolutional layer introduces boundary effects.

Unlike  \eqref{eq:mse_inf}, the problem in \eqref{eq:mse_window} can be readily computed numerically because the inputs and outputs have finite support. Therefore optimizing the set of parameters \(\calH\) with respect to \(\Lfin\) is possible in practice using stochastic gradient descent, as shown in Sections \ref{sec:mtt} and \ref{sec:mid}. The window sizes \(\inW\) and \(\outW\) can be chosen to maximize training performance. For example, one might want the signals to be wide enough to saturate GPU computing while being small enough to fit into memory.

To summarize, we want to find a model \( \bm\Phi(X; \calH) \) that approximates a map from \(X\) to \(Y\), described by the loss function \( \Linf(\calH) \) in \eqref{eq:mse_inf}. However, minimizing \( \Linf(\calH) \) is challenging, especially when the signals involved are of large scale (very wide). Instead, we propose to train the CNN model using \( \Lfin(\calH) \), which is easy to compute. Our proposed approach is theoretically justified by our novel theoretical analysis that provides an upper bound on \( \Linf \) in terms of \( \Lfin \), i.e.,
\begin{equation}\label{eq:bound_form}
    \Linf(\hat\calH) \le \Lfin(\hat\calH) + C.
\end{equation}
To derive this bound, let \(\hat\calH\) be a set of parameters with associated windowed loss \( \Lfin(\hat\calH) \), which can be obtained by minimizing \eqref{eq:mse_window} or other training procedure. Given the following assumptions, we show that \(\hat\calH\) performs almost as well on signals with infinite support -- in terms of \eqref{eq:mse_inf}.

\begin{assumption}\label{assume:jointly_stationary}
    The random signals \(X\) and \(Y\) are jointly stationary following Definition \ref{def:jointly_stationary}.
\end{assumption}
\noindent In many important application domains, signals are at least approximately stationary, which makes our result directly applicable to these cases. Such a result also offers theoretical insights into the fundamental understanding and explainability of CNNs. 
\begin{assumption}\label{assume:model_cnn}
    The model \( \bm\Phi(X; \calH) \) is a convolutional neural network as defined by \eqref{eq:cnn} with a set of parameters \( \calH \).
\end{assumption}
\noindent Given that the signals under consideration exhibit stationarity, it is appropriate to use a convolutional model, which is shift equivariant and is able to exploit stationarity. Assumptions \ref{assume:jointly_stationary} and \ref{assume:model_cnn} are the two keys to our analysis. The remaining assumptions are very mild and typical for model analysis.

\begin{assumption}\label{assume:processes_bounded}
    The stochastic processes \(X\) and \(Y\) are bounded so that \( |X(\bft)| < \infty \) and \( |Y(\bft)| < \infty \) for all \(\bft \in \reals^d\).
\end{assumption}
\noindent Boundedness of the signals is a sufficient condition for the existence and finiteness of some limits. The magnitude of the bound is inconsequential to the final result.

\begin{assumption}\label{assume:finite_filters}
    The filter functions \( h_l \in \calH \) are continuous with a finite width \(K\). That is \(h_l(\bft) = 0\) for all \(\bft \notin [-K/2, K/2]^d\).
\end{assumption}
\begin{assumption}\label{assume:filters_bounded}
    The filters have finite L1 norms \(|| h_l ||_1 \le \infty\) for all \( h_l \in \calH \).
\end{assumption}
\noindent Assumptions \ref{assume:finite_filters} and \ref{assume:filters_bounded} satisfied in typical CNN implementations since the CNN filters are finite in practice.

\begin{assumption}\label{assume:lipschitz}
    The nonlinearities \( \sigma(\cdot) \) are normalized Lipshitz continuous, so that the Lipshitz constant is equal to 1. Mathematically this means that \( | \sigma(x) - \sigma(y)  | \le | x - y | \) for any \( x,y \in \reals \).
\end{assumption}
\noindent Note that the majority of pointwise nonlinear functions used in deep learning, e.g., ReLU, Leaky ReLU, hyperbolic tangent, are normalized Lipshitz for numerical stability. 

Under these assumptions, Theorem \ref{thm:stationary_bound} provides a bound on the generalization error \( \Linf \) of a CNN trained on a finite window and tested on an infinite window. 
\begin{theorem}\label{thm:stationary_bound}
    Let \(\hat\calH\) be a set of filters that achieves a cost of \(\Lfin(\hat\calH)\) on the windowed problem as defined by \eqref{eq:mse_window} with an input window of width \(\inW\) and an output window width \( \outW \). Under Assumptions 1-6, the associated cost \( \Linf(\hat\calH) \) on the original problem as defined by \eqref{eq:mse_inf} is bounded by the following.
    \begin{align}
        \Linf(\hat\calH) & \le \Lfin(\hat\calH) + \E{X(0)^2} C + \sqrt{\Lfin(\hat\calH) \E{X(0)^2} C} \label{eq:stationary_bound}\\
        C &= \frac{H^2}{\outW^d} \max\left(0, [\outW + LK]^d - \inW^d\right)
    \end{align}
    In \eqref{eq:stationary_bound}, \(H = \prod_l^L || h_l ||_1\) is the product of all the L1 norms of the CNNs filters, \(L\) is the number of layers, and \(K\) is the width of the filters.
\end{theorem}
The proof is provided in Appendix \ref{appendix:stationary_bound}. The derivation exploits shift-equivariance of CNNs, which preserve joint stationarity of the signals. Theorem \ref*{thm:stationary_bound} shows that minimizing \( \Linf \) can be approximated by instead minizing \( \Lfin \). The maximum deterioration in performance is bounded by a quantity that is affected by the variance of the input signal, number of layers, filter widths, and the size of the input and output windows. The inequality in \eqref{eq:stationary_bound} reduces to \(\Linf(\hat\calH) \le \Lfin(\hat\calH)\) whenever \( \inW \ge \outW + LK \). In this special case, the output becomes unaffected by padding, which explains the last two terms on the right-hand side of \eqref{eq:stationary_bound}.

%% file: journal/representing.tex
% !TeX root = journal.tex

\section{Representing Sparse Problems as Mixtures}\label{sec:representing}

CNNs are well suited to processing images or, more generally, multi-dimensional signals.
In various spatial problems \cite{Mox20-MobileWirelessNetwork, Mox22-Learning, Owerko23-MultiTargetTracking}, the decision variables are finite sets of real vectors with random cardinality such as \( \calX = \{ \bfx \in \reals^d \} \).
Therefore, in this section we propose a method to represent sets of vectors such as \( \calX \) by multi-dimensional signals \({X(\bft; \calX): \reals^d \mapsto \reals}\).

\begin{figure*}
    \centering
    \includegraphics[width=\figWidth\linewidth]{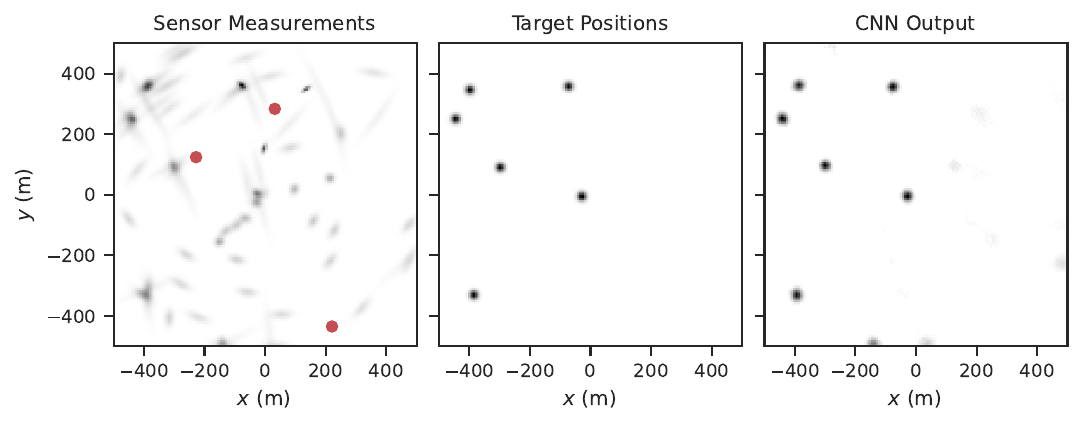}
    \caption{Example 2D signals from multi-target tracking; see Section \ref{sec:mtt} for a full problem description.
        \emph{(Sensor Measurements)} \( \bar\bfZ_n \) represents the measurements made by the sensors in one time-step -- their positions are marked in red. The sensors are modeled as range-bearing with additive noise, therefore in Cartesian coordinates, the distribution has a distinct banana shape. This example was cherrypicked to show multiple sensors in the 1km\textsuperscript{2} area, but in simulations the sensors can be up to \(R\) meters outside this area.
        \emph{(Target Positions)}: \( \bfX_n \) represents the ground truth positions of all the targets.
        \emph{(CNN Output)}: \( \bm\Phi(\bar\bfZ_n; \hat\calH) \) is the output of the trained CNN. The CNN is trained to approximate the target positions image, given the past \(K\) sensor measurement images.
    }
    \label{fig:mtt_images}
\end{figure*}

Representing the sets by a superposition of Gaussian pulses is an intuitive approach. Specifically,
\begin{equation}\label{eq:intensity_gaussian}
    X(\bft; \calX) \defeq \sum_{\bfx \in \calX} (2 \pi \sigma_X^2)^{-1}
    \exp(-\frac{1}{2\sigma_X^2}||\bft-\bfx||^2_2).
\end{equation}
where there is one pulse with variance \( \sigma_X^2 \) for each element in the set. The second image in Figure \ref{fig:mtt_images} exemplifies this.

This approach can be made more general by incorporating additional information about the distribution of the set.
Assume that each element \(\bfx \in \calX\) is in fact a sample from a distribution that depends on \(\bft\).
The density function \(g(\bfx\mid\bft)\) describes this dependence.
For example, when \( \bfx \) is a noisy measurement of some underlying quantity \(g(\bfx\mid\bft)\) describes the likelihood of sampling \( \bfx \) given \( \bft \) as the ground truth.
There could exist multiple sensors making measurements, each with a different density function \(g_i(\bfx\mid\bft)\).
Denote \( x_i \in \calX \) to make it clear that each element in \(\calX\) is associated with some density function \(g_i\).
Therefore the intensity function \( X(\bft; \calX) \) can be written as a sum over these densities:
\begin{equation}\label{eq:intensity_density}
    X(\bft; \calX) \defeq \sum_{\bfx_i \in \calX} g_i(\bfx_i | \bft).
\end{equation}
The function \(g_i\) is sometimes known as a \emph{measurement model}. In the first image in Figure \ref{fig:mtt_images}, the measurements contain noise applied in polar coordinates around sensors. This results in an arc-shaped distribution. Notice that \eqref{eq:intensity_gaussian} is a special case \eqref{eq:intensity_density} when the distribution is Gaussian.

In practice, we window and discretize the intensity function \( X(\bft; \calX) \). The function is windowed to width \(T\) around zero and sampled with resolution \( \rho \). This produces a \(d\)-dimensional tensor \( \bfX \in \reals^{N^d} \) with width \( N = \lfloor \rho T \rfloor \) along each dimension. Similarly, given a set \( \calY = \{ \bfy \in \reals^d \} \) we construct a tensor \( \bfY \in \reals^{N^d} \). This allows training a CNN to learn a mapping between \(\bfX\) and \(\bfY\).

To summarize, our original problem was to model the relationship between \(\calX\) and \(\calY\). These are random sets and we assume that there exists a dataset of their realizations. We can represent each pair in this dataset as a tensor \(\bfX\) and \(\bfY\), respectively. Therefore, we can use a CNN to learn a map between these tensors. Hopefully, we find parameters so that \( \hat\bfY := \bm\Phi(\bfX; \calH) \) is close to \(\bfY\). However, we also need a method to recover an estimate of \(\calY\) from \(\hat\bfY\), the output of our CNN.

To achieve this, we propose the following approach. Assume that \(\bfY\) is obtained by following \eqref{eq:intensity_gaussian} and discretizing; this is the case in our experiments. Notice that the L1 norm of the output image \(||Y(\bft; \calY)||_1 = |\calY|\) is by construction euqal to the cardinality of \(\calY\). If during discretization we preserve the L1 norm, then \(\bfY\) has a similar property for all elements of \(\calY\) within the window. If CNN training is successful then \(\hat\bfY\) will retain this property. Since we know the number of Gaussian components, we can use k-means clustering or fit a Gaussian mixture via expectation maximization. In our implementation, we use the subroutines provided by Scikit-Learn \cite{Pedregosa11-Scikitlearn}.

%% file: journal/architecture.tex
% !TeX root = journal.tex
\section{Architecture}

Theorem \ref{thm:stationary_bound} suggests that training a CNN on small, representative windows of multi-dimensional input-output signals is effective at learning a mapping between the original signal pairs. Such as CNN can later generalize to arbitrary scales by leveraging the shift equivariance property of CNNs.
% Note that the theorem does not depend on the dimensionality of the parameter \(\bft\). Since the nonlinearities are pointwise, Theorem \ref{thm:stationary_bound} is applicable to \(d\)-dimensional signals. 
In this section, we describe a fully convolutional architecture for the processing of multi-dimensional signals.

\subsection{Fully convolutional encoder-decoder}

\begin{figure}[ht]
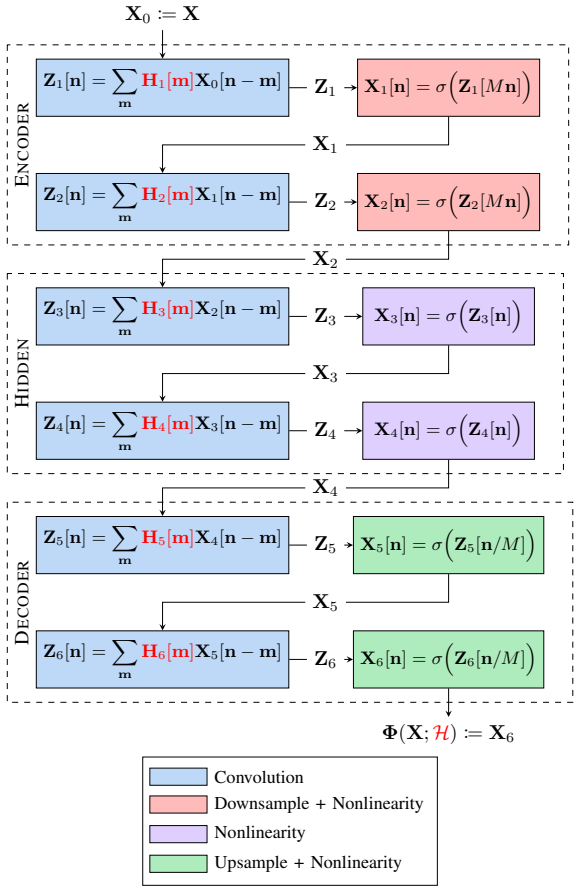

    \centering
    \includestandalone[width=\figWidth\linewidth]{figures/architecture}
    \caption{Diagram of the proposed CNN architecture. Above we depict a CNN with two encoder, hidden, and decoder layers. This illustrates the architecture but is different from the configuration used in the experiments.}
    \label{fig:encoder_decoder}
\end{figure}

The inputs and outputs of the architecture are multidimensional signals:
\[\begin{split}
        \bfX \in \underbrace{\reals^N \times \dots \times \reals^N}_{d~\text{times}} \times \reals^{F_0} \\
        \text{and}\quad
        \bm\Phi(\bfX, \calH) \in \underbrace{\reals^N \times \dots \times \reals^N}_{d~\text{times}} \times \reals^{F_L}
    \end{split}\]
with \(F_0\) and \(F_L\) features, respectivelly. The first \(d\) dimensions are the translationally symmetric space where the task takes place. To process these signals we propose a fully convolutional encoder-decoder architecture as shown in Figure \ref{fig:encoder_decoder}. The proposed architecture is organized into three sequential components: the encoder, the hidden block, and the decoder. Each of these components contains several layers with \(L\) in total.
The input to the first layer is \(\bfX\) and the output of the final layer is \(\bm\Phi(\bfX, \calH)\).
Each layer contains a convolution, a pointwise nonlinearity, and possibly a resampling operation.

Convolution operations are the first component of every layer. The input to the \(l^\text{th}\) layer is a tensor \( {\bfX_{l-1}} \) of order $d+1$, such that \( {\bfX_{l-1}[\bfn] \in \reals^{ F_{l-1} }} \) with \(F_{l-1}\) features and index \(\bfn \in \mathbb{Z}^d\).
This input is passed through a multiple-input multiple-output convolutional filter
\begin{equation}\label{eq:multi_convolution}
    \bfZ_l[\bfn] = \sum_{\bfm\in\bbZ^d} \bfH_{l}[\bfm] \bfX_{l-1}[\bfn - \bfm],
\end{equation}
to obtain an intermediate output \({ \bfZ_{l}[\bfn] \in \reals^{ F_l } }\) with \(F_l\) features. In \eqref{eq:multi_convolution}, \(\bfH_l\) is a high-order tensor with \({ \bfH_l[\bfn] \in \reals^{F_l \times F_{l-1}} }\) representing a filter with a finite width \(K\). As a result, the filter output \({ \bfZ_{l} }\) is also a high-order tensor with finite width that is indexed by \(\bfn \in \mathbb{Z}^d\).

The encoder and decoder are mirror images of each other. The encoder downsamples the signal, while the decoder upsamples it. This effectively compresses the signal, so that the hidden block can efficiently process it. Each encoder layer is composed of a convolutional layer, as defined in \eqref{eq:multi_convolution}, followed by a downsampling operation and a pointwise nonlinearity:
\begin{equation}\label{eq:downsampling}
    \bfX_l[\bfn] = \sigma \left( \bfZ_l[M\bfn] \right).
\end{equation}
Each layer in the encoder downsamples the input by a factor of \(M\). Conversely, the decoder upsamples its inputs by the same factor. Much like an encoder layer, each decoder layer contains a convolution followed by an upsampling operation and a pointwise nonlinearity:
\begin{equation}\label{eq:upsampling}
    \bfX_l[\bfn] = \sigma \left( \bfZ_l[\bfn/M] \right).
\end{equation}
In \eqref{eq:upsampling}, the use of square brackets in \( \bfZ_l[\bfn/M] \) indicates that the index is rounded to the nearest integer.

The hidden block layers are interposed between the encoder and decoder. Each hidden layer is composed of a convolution followed by a pointwise nonlinearity:
\begin{equation}\label{eq:nonlinearity}
    \bfX_l[\bfn] = \sigma \left( \bfZ_l[\bfn] \right).
\end{equation}
They are similar to the encoder and decoder layers, but they do not contain any resampling operations.

The proposed encoder-decoder architecture has two main benefits. First, it is compatible with Theorem \ref{thm:stationary_bound} because it is fully convolutional. Therefore, it preserves the stationarity of the input signals \cite{Zou20-LipschitzBounds}. Second, it allows for efficient processing of multi-dimensional signals.
The encoder downsamples the input signal, reducing the amount of computation needed in the hidden block.
Afterwards, the decoder upsamples the processed signal back to the original representation.
This compression is especially beneficial for our experiments because, as can be seen from figure \ref{fig:mtt_images}, the tasks we consider have sparse input and output signals.

% Then we can learn a fully convolutional model as follows, 
% \begin{equation}\label{eq:mse_image}
%     \min_{\calH} \E{ \frac{1}{N^2} || \bm\Phi(\bfX; \calH) - \bfY ||^2_2 }.
% \end{equation}
% Notice that \eqref{eq:mse_image} is the discrete equivalent of \eqref{eq:mse_window}. In particular, sampling $\sqcap_\inW X$ and $\sqcap_B Y$ at regular intervals To make this parallel closer we assume that there is a constant \( \rho \) that relates the window size \(T\) with \(N\) according to \( N = \lfloor \rho T \rfloor \). In this context, \(\rho\) is the spatial resolution of the CNN. To apply Theorem \ref{thm:stationary_bound} this constant should be fixed while changing the window size \(T\). 

\subsection{Hyperparameters}\label{subsec:mtt_hyperparameters}

In practice, we found the following hyperparameter values to be effective in our simulations. Our encoder and decoder layers have filters with width \(K = 9\), \(F_l = 128\) channels, and a resampling factor of \(M = 2\). Meanwhile, each hidden layer uses filters with width \(K=1\) and \(F_l = 1024\) channels. We use three encoder layers, four hidden layers, and three decoder layers. In Section \ref{sec:mtt} we will elaborate on how we arrived at these hyperparameters.

There are several possible practical methods of downsampling and upsampling in the encoder and decoder blocks. In our best-performing variant of the architecture, we combine them with convolutions. In the encoder, we use a convolution with a stride of two. Meanwhile, in the decoder, we use transpose convolutions with a stride of two. This approach is used in \cite{Rombach22-HighResolution}, although it is prone to producing checkerboard artifacts \cite{Odena16-Deconvolution}. Alternatively, the downsampling can be performed separately or using a pooling layer such as average pooling. Similarly, the transpose convolutions can be replaced by an upsampling or interpolation layer followed by a convolution. However, in our simulations, we found this approach to be less performant, although it allowed training to converge slightly faster.

%% file: figures/architecture.tex
\begin{tikzpicture}
    \pgfdeclarelayer{bg}    % declare background layer
    \pgfsetlayers{bg, main} % set layer order (main is the standard layer)

    % Control separation between blocks
    \def \deltainput     {(0.0, 0.5)}
    \def \deltaoutput    {(0.0,-0.5)}
    \def \deltalayer     {2.0}
    \def \xlabel         {2.85}
    \def \deltasigma     {(5.0, 0.0)}
    \def \deltapad       {-0.5, 0.75}

    % compute y position of Lth layer
    \newcommand{\ylayer}[1]{-#1*\deltalayer}

    % Definition of a block for block diagrams
    \tikzstyle{block} = [
    rectangle,
    minimum width = 3.0cm,
    minimum height = 1.0cm,
    font = \small,
    ]

    \tikzstyle{wide} = [
    block,
    minimum width = 4.3cm
    ]

    \tikzstyle{blockstyle} = [
    fill opacity = 0.7,
    text opacity = 1.0,
    draw = black,
    text = black
    ]

    \tikzstyle{boundingbox} = [
    dashed
    ]

    \tikzstyle{conv} = [blockstyle, fill = pastel0]
    \tikzstyle{convlocal} = [blockstyle, fill = pastel1]
    \tikzstyle{downsample} = [blockstyle, fill = pastel3]
    \tikzstyle{sigma} = [blockstyle, fill = pastel4]
    \tikzstyle{upsample} = [blockstyle, fill = pastel2]

    % Text displayed within different kinds of blocks
    \newcommand{\layerconv}[1]{$\displaystyle{
                \bfZ_{#1}[\bfn] = \sum_{\bfm} \red{\bfH_{#1}[\bfm]} \bfX_{\pgfinteval{#1-1}}[\bfn-\bfm]
            }$}
    \newcommand{\layerconvlocal}[1]{$\displaystyle{
                \bfZ_{#1}[\bfn] = \red{\bfH_{#1}} \bfX_{\pgfinteval{#1-1}}[\bfn]
            }$}
    \newcommand{\layerdownsample}[1]{$\displaystyle{
                \bfX_{#1}[\bfn] = \sigma \Big( \bfZ_{#1}[M\bfn] \Big)
            }$}
    \newcommand{\layersigma}[1]{$\displaystyle{
                \bfX_{#1}[\bfn] = \sigma \Big( \bfZ_{#1}[\bfn] \Big)
            }$}
    \newcommand{\layerupsample}[1]{$\displaystyle{
                \bfX_{#1}[\bfn] = \sigma \Big( \bfZ_{#1}[\bfn/M] \Big)
            }$}

    % Encoder Layers
    \foreach \L in {1,2}
        {
            %
            % Draw Connectors between layers
            \path (0, \ylayer{\L}) node [wide, conv] (Filter\L) {\layerconv{\L}};
            %
            % Draw Nonlinearity Block
            \path (Filter\L) + \deltasigma node [block, downsample] (Sigma\L) {\layerdownsample{\L}};
        }
    % Hidden Layers
    \foreach \L in {3,4}
        {
            %
            % Draw Connectors between layers
            \path (0, \ylayer{\L}) node [wide, conv] (Filter\L) {\layerconv{\L}};
            %
            % Draw Nonlinearity Block
            \path (Filter\L) + \deltasigma node [block, sigma] (Sigma\L) {\layersigma{\L}};
        }
    % Decoder Layers
    \foreach \L in {5,6}
        {
            % Draw Connectors between layers
            \path (0, \ylayer{\L}) node [wide, conv] (Filter\L) {\layerconv{\L}};
            %
            % Draw Nonlinearity Block
            \path (Filter\L) + \deltasigma node [block, upsample] (Sigma\L) {\layerupsample{\L}};
        }
    % Draw connectors inside of layers
    \foreach \L in {1,...,6}
        {
            \node (Z\L) at (\xlabel, \ylayer{\L}) {$\bfZ_\L$};
            \draw [-stealth] (Filter\L.east)
            -- (Z\L)
            -- (Sigma\L.west);
        }

    % Connectors between layers
    \foreach \L in {1,...,5}
        {
            % \Lplus = \L + 1
            \pgfmathtruncatemacro{\Lplus}{\L+1}
            % \ymid is the midpoint between two layer heights
            \pgfmathsetmacro{\ymid}{\ylayer{\L}/2+\ylayer{\Lplus}/2}

            \node (Aux\L) at (\xlabel, \ymid) {$\bfX_\L$};
            \draw[-stealth] (Sigma\L.south)
            |- (Aux\L)
            -| (Filter\Lplus.north);
        }

    % Legend
    \matrix
    [
        draw,
        anchor=north,
        column 1/.style={block, minimum height = 0.3cm, minimum width = 1cm},
    ]
    (legend) at ($ .5*(Filter6.west) + .5*(Sigma6.east) + (0,-1.7) $)
    {
        \node[conv, label=right:Convolution] {};                      \\
        % \node[convlocal, label=right:Pointwise Convolution] {};       \\
        \node[downsample, label=right:Downsample + Nonlinearity] {}; \\
        \node[sigma, label=right:Nonlinearity] {};                    \\
        \node[upsample, label=right:Upsample + Nonlinearity]{};      \\
    };

    % Model input
    \draw[stealth-] (Filter1.north)
    -- ++\deltainput node[anchor=south] {$\bfX_0 \defeq \bfX$};

    % Model output
    \draw[-stealth] (Sigma6.south)
    -- ++\deltaoutput node[anchor=north] {\( \bm\Phi(\bfX; \red{\calH}) \defeq \bfX_6 \)};

    % Light shading to signify layer groups. Layer 1   
    \begin{pgfonlayer}{bg}
        \draw[boundingbox] ($ (Filter1.west) + (\deltapad) $) rectangle ($ (Sigma2.east)  - (\deltapad) $);
        \node[rotate=90, above] at ($ 0.5*(Filter1.west) + 0.5*(Filter2.west) $) {\textsc{Encoder}};

        \draw[boundingbox] ($ (Filter3.west) + (\deltapad) $) rectangle ($ (Sigma4.east)  - (\deltapad) $);
        \node[rotate=90, above] at ($ 0.5*(Filter3.west) + 0.5*(Filter4.west) $) {\textsc{Hidden}};

        \draw[boundingbox] ($ (Filter5.west) + (\deltapad) $) rectangle ($ (Sigma6.east)  - (\deltapad) $);
        \node[rotate=90, above] at ($ 0.5*(Filter5.west) + 0.5*(Filter6.west) $) {\textsc{Decoder}};
    \end{pgfonlayer}
\end{tikzpicture}

%% file: journal/mtt.tex
% !TeX root = journal.tex

\section{Multi Target Tracking}\label{sec:mtt}

Multi-target tracking (MTT) is a type of sequential estimation problem where the goal is to jointly determine the number of objects and their states from noisy, discrete-time sensor returns.
Such sensor returns can include the pulse energy observed across a series of sampled time delays in radar systems or the pixel intensities observed from a camera system.
It is distinct from \emph{state filtering} in that object appearance and disappearance (i.e., birth and death) must be addressed in the problem formulation.
We direct the reader to \cite{Vo15-MultitargetTracking} for a detailed overview of the field and the relevant application areas.

We focus on the canonical application known as \emph{point object tracking} where sensor returns are quantized via a detection process into measurements before the tracker is applied.
To make the problem definition more precise, we introduce notation for the multi-target state and measurement sets at the \(n^\text{th}\) time-step.
Let the \emph{multi-target state} be denoted by the set, \(\calX_n\). Each element of this set represents the state of a target in some space.
At each time step each sensor will produce noisy measurements of the targets, resulting in a set \(\calZ_n\) of measurements in a different space.
The goal in MTT is to estimate the multi-target state \(\calX_n\) given all the preceding measurements \(\calZ_1,...,\calZ_n\).
The principal challenge in this task is \emph{measurement origin uncertainty}; it is not known \emph{a priori} which measurements were generated by true targets, were the result of an independent clutter process (false positive) or did not generate measurements (false negative).
A direct Bayesian solution to the measurement origin uncertainty problem with multiple sensors requires solving an NP-hard multi-dimensional ranked assignment problem \cite{Vo19-MultiSensorMultiObject}.

Algorithms in this area are well-studied and include global nearest neighbor techniques~\cite{Blackman99}, joint probabilistic data association~\cite{Barshalom09-JPDAF}, multi-hypothesis tracking~\cite{Blackman04-Multiple}, and belief propagation~\cite{Meyer18-BPTracking}.
Recently random finite sets (RFS) have emerged as a promising all-in-one Bayesian framework that addresses the measurement origin uncertainty problem jointly~\cite{Mahler07-Statistical, Mahler14-AdvancesStatistical}.
In particular, \emph{labeled RFS} were introduced in \cite{Vo13-LabeledRandomFinite} specifically for MTT, resulting in the Generalized Labeled Multi-Bernoulli (GLMB) \cite{Vo14-LabeledRandomFinite, Vo17-EfficientImplementation} and Labeled Multi-Bernoulli (LMB) \cite{Reuter2014, Reuter2017} filters.
However, the GLMB and LMB filters exhibit exponential complexity in the number of sensors \cite{Vo17-EfficientImplementation, Vo19-MultiSensorMultiObject, Beard20-SolutionLargeScale}.

Our proposed analytical and computational framework is an excellent fit to overcome these scalability issues for two reasons: (i) it is independent of the number of sensors and targets within a specific region and (ii) it can leverage transfer learning to perform MTT in large areas. The proposed CNN tracker is trained on a small 1km\textsuperscript{2} area and tested on larger areas up to 25km\textsuperscript{2} where the number of sensors and targets are increased proportionally to maintain a fixed density of each.
The computational complexity of this approach is proportional to the simulation area rather than the number of targets and sensors.
Based on Theorem \ref{thm:stationary_bound}, we would expect the tracking performance from training to be maintained as this scaling up is carried out.

\subsection{Modeling targets and sensors}\label{subsec:mtt_modeling}

The multi-target state changes over time. First, new targets may be born according to a Poisson distribution with mean \(\lambda_\text{birth}\).
Second, old targets may disappear with probability \(p_\text{death}\). Finally, the existing target's state evolves over time.
In our simulations we use a (nearly) constant velocity (CV) model \cite{Baisa20-DerivationConstant, RongLi03-SurveyManeuvering}.
In the CV model the state \(\bfx_{n,i} \in \calX\) at time-step \(n\) of the \(i^\text{th}\) target is given by \eqref{eq:state}.
\begin{equation}\label{eq:state}
    \bfx_{n,i} = \bmat{\bfp_{n,i} & \bfv_{n,i}}
\end{equation}
A target state is composed of its position \(\bfp_{n,i} \in \reals^2\) and velocity \(\bfv_{n,i} \in \reals^2\). The state evolves according to \eqref{eq:cv}.
\begin{equation}\label{eq:cv}
    \begin{split}
        \Delta \bfp_{n,i} & = \tau \dot\bfp_{n,i} + \frac{\tau^2}{2} \bm\mu_{n,i} \\
        \Delta \bfv_{n,i} & = \tau \bm\mu_{n,i}                                  \\
    \end{split}
\end{equation}
The CV model is discretized with time-step \( \tau \) seconds. Also, note that the acceleration \( \bm\mu_{n,i} \in \reals^2 \) is sampled from a multivariate normal distribution, 
\begin{equation}
    \bm\mu_{n,i} \sim\calN \left(0, \bsmat{\sigma_a^2 & 0 \\ 0 & \sigma_a^2} \right).
\end{equation}

There are \(M\) sensors that provide measurements of the multi-target state. Consider the \(j^\text{th}\) sensor with position \(\bfs_j \in \reals^2\). If a target with state \(\bfx_{n,i} \in \calX_n\) is within the sensor's radial range of \(R\) km, then at each time step the sensor may detect the target with probability \(p_\text{detect}\). When this occurs then a corresponding measurement \(\bfz_{n,i}^j\) is added to the measurement set \(\calZ_n\). Multiple sensors can detect the same target, but each can provide at most one detection per target. The measurement \(\bfz_{n,i}^j\) is stochastic and its distribution is known as the \emph{measurement model}. In our simulations, the measurement model is range-bearing relative to the position of the \(j^\text{th}\) sensor:
\begin{equation}\label{eq:measurement_model}
    \begin{split}
        \bfz_{n,i}^j &\sim \calN \left(
        \bmat{|| \bfp_{n,i} - \bfs_j || \\ \angle ( \bfp_{n,i} - \bfs_j )},
        \bmat{\eta_r^2 & 0 \\ 0 & \eta_\theta^2}
        \right ).
    \end{split}
\end{equation}
The measurements are Gaussian distributed with a mean equal to the distance and bearing to the target. The covariance matrix is diagonal with a standard deviation of \(\eta_r\) meters in range and \(\eta_\theta\) radians in angle. Additionally, clutter is added to the measurement set \(\calZ_n\). The number of clutter measurements is Poisson with mean \(\lambda_\text{clutter}\) per sensor. Clutter is generated independently for each sensor, with clutter positions sampled uniformly within its range.

\subsection{Simulations}

To train the CNN we construct a dataset of images representing the MTT problem within a \(T = 1\)km wide square window. Similarly, we construct a testing dataset for window widths between 1km and 5km. 
For any particular window of width \(T\), constructing a dataset of images for training or testing is the same. First, we simulate the temporal evolution of the multi-target state \(\calX_n\) and at each time step generate sensor measurements \( \calZ_n \). Table \ref{tab:mtt_parameters} describes the parameters used for the simulations. At the beginning of each simulation we initialize \(\Poisson(\lambda_\text{initial})\) targets and \(\Poisson(\lambda_\text{sensor})\) sensors. Their positions are uniformly distributed within the simulation region, while the initial velocity for each target is sampled from \(\calN(0, \sigma_\text{v}^2)\).

\begin{table}
    \centering
    \caption{Parameter values used for MTT simulations. Temporal quantities are defined per time step. }
    \label{tab:mtt_parameters}
    \begin{tabular}{lll}
        \toprule
        Parameter                    & Value                  & Description                        \\
        \midrule
        \( p_\text{death} \)         & 0.05                   & Target death probability.          \\
        \( p_\text{detect} \)        & 0.95                   & Target deteciton probability.      \\
        \( \sigma_V \)               & 10 m                   & Target image kernel width.         \\
        \( \sigma_\text{a} \)        & \( 1 \tau^2\) m$^2$/s  & CV model acceleration noise.       \\
        \( \sigma_\text{v} \)        & \( 5.0 \tau^2\) m/s    & Initial target velocity noise.     \\
        \( \eta_r \)                 & 10 m                   & Range measurement noise.           \\
        \( \eta_\theta \)            & 0.035 rad              & Bearing measurement noise.         \\
        \( \lambda_\text{birth} \)   & \( 0.5 T^2\tau \)      & Mean number of targets born.       \\
        \( \lambda_\text{initial} \) & \(10 T^2\)             & Mean initial number of targets.    \\
        \( \lambda_\text{sensor} \)  & \( 0.25 T^2 \)         & Mean number of sensors.            \\
        \( \lambda_\text{clutter} \) & \( 40\tau \)           & Mean clutter rate per sensor.      \\
        \( R \)                      & 2 km                   & Sensor range.                      \\
        \( \tau \)                   & 1 s                    & Time-step duration.                \\
        \( \rho \)                   & \(\frac{1000}{128}\) m & Image spatial sampling resolution. \\
        \bottomrule
    \end{tabular}
\end{table}

Notice that sensors outside of the window \(T\) can make measurements of targets within the window. Unless we want to break our assumption of stationarity, we have to include these measurements as well. Thinking back to equation \eqref{eq:mse_inf}, the images should be a cropped version of some infinitely wide signal. Hence, it is actually necessary to simulate the targets and sensors over a window larger than \(T\). Luckily, the sensors have a finite range \(R\)km, so we run simulations over a square window of width \( 2R + T \) -- padding the simulation area by \(R\)km on each side. Similarly, we also keep track of targets that could be born outside \(T\)km window and targets that leave the window but might return.

To generate training data, we ran 10,000 simulations with 100 steps each at a window width of \(T = 1\)km. We generate images \( \bfZ(\calZ_n) \) and \(\bfX(\calX_n) \) from \(\calZ_n\) and \(\calX_n\), respectivelly, using the procedure outlined in Section \ref{sec:representing}. In this section, \(\bfX\) denotes the \emph{output} of the CNN for consistency with the literature. To provide temporal information, the input to the CNN is actually a stack of \(K = 20\) past sensor measurement images:
\begin{equation}\label{eq:input_stack}
    \bar\bfZ_n \defeq \bmat{\bfZ(\calZ_{n-K+1}) & \dots & \bfZ(\calZ_{n})}.
\end{equation}
Data for \(n < K\) is excluded. We train the model for 84 epochs using AdamW \cite{Loshchilov22-DecoupledWeight} with a batch size of 32, a learning rate of \(6.112 \times 10^{-6}\), weight decay of \( 0.07490 \). We arrived at these hyperparameters and the ones in Subsection \ref{subsec:mtt_hyperparameters} by running a random search using the Optuna library \cite{Akiba19-OptunaNextgeneration}. The hyperparameters were sampled using the built-in tree-structured parzen estimator \cite{Bergstra11-Algorithms}.

\subsection{Baseline Approaches}

Our baseline for comparing the CNN's performance is defined by two state-of-the-art filters.
The filters used are the sequential Monte Carlo multi-sensor iterated corrector formulation of the Labeled Multi-Bernoulli (LMB) and the $\delta$-Generalized Labeled Multi-Bernoulli (GLMB) \cite{Papi16-IteratedCorrectorGLMB}.
The bearing-range measurement models, sensor generation, and dynamic model for the targets are the same as described in Subsection \ref{subsec:mtt_modeling}.

Both filters used the Monte Carlo approximation for the adaptive birth procedure from \cite{Trezza22-MultisensorAdaptiveBirth}.
The proposal distribution is created by decomposing the state space into observable ($x_o$) and unobservable ($x_u$) states, with observable ($p^o_B(x_o, l_+)$) and unobservable ($p^u_B(x_u, l_+)$) prior densities.
For the observable states, $\mathbb{X}_o$, an uninformative uniform prior distribution $p^o_{B}(x_o, l_+) = \mathcal{U}(\mathbb{X}_o)$ was used.
The unobservable states, which for this scenario are the velocities, were sampled from a zero-mean Gaussian distribution:
\begin{equation}
    p^u_{B}(x_u, l_+) = \mathcal{N}(x_u; 0, \begin{bsmallmatrix}\sigma_\text{v}^2 & 0\\ 0 & \sigma_\text{v}^2\end{bsmallmatrix}).
\end{equation}
As the scenario scaled up in the number of targets and sensors, the number of measurements increased significantly.
To accommodate for this, the number of Gibbs iterations was increased for each scenario.
Due to computational limitations, window sizes of \(T \in \{1,2,3\} \)km were used, with 2000, 4000, and 6000 Gibbs iterations, respectively.
Every 10 Gibbs iterations the current solution was reset to the all miss-detected tuple, to encourage exploration.

\begin{table}
    \centering
    \caption{$\delta$-GLMB and LMB parameters.}
    \label{tab:rfs_lmb_parameters}
    \begin{tabular}{lll}
        \toprule
        Parameter            & Value            & Description                      \\
        \midrule
        \( S_\text{birth} \) & 2000, 4000, 6000 & Birth Gibbs iterations.          \\
        \( r_{B, max} \)     & 1.0              & Maximum birth probability.       \\
        \( \lambda_{B,+} \)  & 1.0              & Expected birth rate.             \\
        \( p_\text{max} \)   & 0.001            & Maximum association probability. \\
        \bottomrule
    \end{tabular}
\end{table}

\subsection{Results}

\begin{figure*}
    \centering
    \includegraphics[width=\figWidth\linewidth]{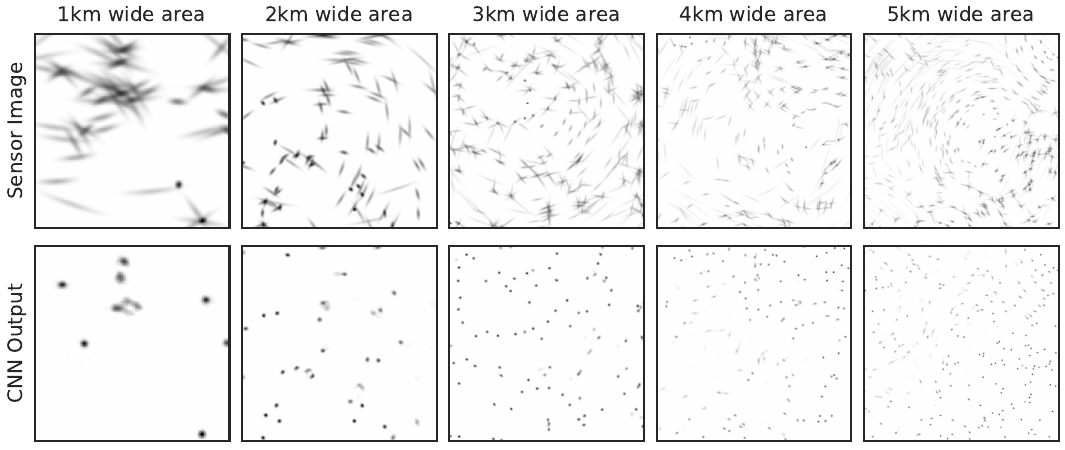}
    \caption{The input and output of the CNN at different window widths of \(T = \{1,...,5\} \)km \cite{Owerko23-MultiTargetTracking}. To increase the contrast on this figure we rescaled the sensor image to \( \log( \bfZ(\calZ_n) + 0.001 ) \), which is shown in the first row. The second row shows the corresponding output of the trained CNN, \( \bm\Phi(\bar\bfZ_n; \calH^*) \), without any adjustments.}
    \label{fig:multiscale}
\end{figure*}

To test the performance of the trained CNN we ran 100 simulations at each window size \(T \in \{1,2,3,4,5\} \)km and constructed images from them. Figure \ref{fig:multiscale} shows example inputs and outputs of the CNN, picked at random during testing at each window size. We compare our results against state-of-the-art filters: Labeled Multi-Bernoulli (LMB) and Generalized Labeled Multi-Bernoulli (GLMB).

The output of the CNN and the output of the LMB and GLMB filters are not in the same space. Specifically, the output of the CNN filter is an image, whereas the LMB and GLMB filters output a set of vectors in \(\calX\) and a set of weights. The vectors represent the estimated states of the targets, while the weights are their existence probability. This makes it difficult to compare the two. To draw a fair comparison, we consider two different metrics: the mean squared error (MSE) and optimal sub-pattern assignment (OSPA) \cite{Schuhmacher08-Consistent}.

\begin{figure}
    \centering
    \includegraphics[width=\figWidth\linewidth]{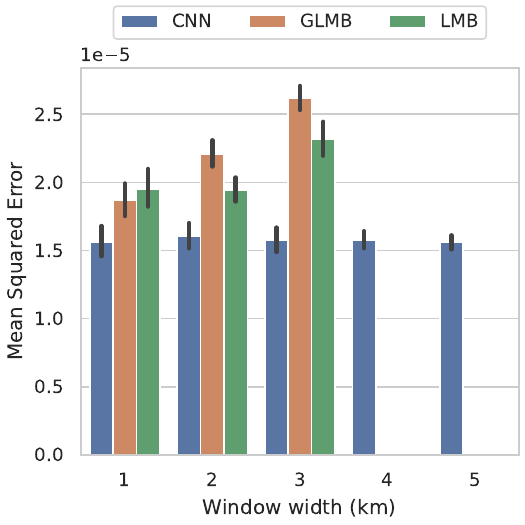}
    \caption{Comparison of MSE of the three filters for different window sizes \(T\). We report the mean value for 100 simulations at each scale. The error bars indicate a 0.95\% confidence interval in the MSE. }
    \label{fig:mse_transfer}
\end{figure}

In this context, the MSE is computed between the CNN output and the image representing the true target positions, as defined by equation \eqref{eq:intensity_gaussian}. We convert the LMB and GLMB outputs to image using a superposition of Gaussians, as described by equation \eqref{eq:intensity_gaussian}. The only notable difference is that we weigh the Gaussians by the estimated existence probability. Figure \ref{fig:mse_transfer} shows that the MSE of the CNN is unaffected as the window size increases. This directly supports Theorem \ref{thm:stationary_bound}. Moreover, this is the case even though our CNN implementation employs padding. Therefore, the practical effect of padding is lower than the second and third terms in \eqref{eq:stationary_bound} suggest.

However, MSE does not fairly compare the different approaches. The output of the MTT problem should be a set of estimates. OSPA \cite{Schuhmacher08-Consistent}, as defined in Definition \ref{def:ospa}, allows us to measure the average ``distance'' between two sets of vectors, possibly of different cardinalities. We use OSPA  with a cutoff distance of \(c = 500\) meters. We describe the method used to extract a set of vectors from the CNN output image in Section \ref{sec:representing}.

\begin{definition}\label{def:ospa}
    (OSPA) If $\calX = \{\bfx_i\}_{i=1,...,n}$ and $\calY = \{\bfy_i\}_{i=1,...,m}$ are two sets of vectors in $\reals^d$ with $m \le n$ then OSPA is defined as
    $$ \text{OSPA}(\calX, \calY, c) =
        \min_{\pi \in \Pi_n} \sqrt{\frac{1}{n} \left[\sum_{i=1}^m ||\bfx_i - \bfy_{\pi_i} ||_2^2 + c(n-m) \right] } $$
    where $c$ is some cutoff distance and $\pi = [\pi_1,...,\pi_n]$ is a permutation.
\end{definition}

\begin{figure}
    \centering
    \includegraphics[width=\figWidth\linewidth]{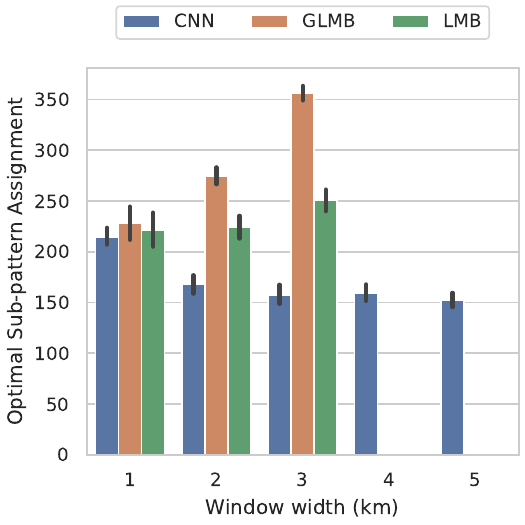}
    \caption{Comparison of OSPA of the three filters for different window sizes \(T\) \cite{Owerko23-MultiTargetTracking}. We report the mean value for 100 simulations at each scale. The error bars indicate a 0.95\% confidence interval in the MSE. }
    \label{fig:ospa_transfer}
\end{figure}

The CNN maintains consistent performance across scales and outperforms the GLMB and LMB filters. This claim is supported by Figure \ref{fig:ospa_transfer}, which shows the OSPA for different values of \(T\) from testing. GLMB and LMB filters have comparable performance to the CNN at \(T = 1\)km, but their performance degrades as the region is scaled up due to truncation issues. In contrast, the performance of the CNN improves at scale. Within a 1km\textsuperscript{2} region the CNN obtains an OSPA of \( 215 \pm 73 \) meters. This decreases to \( 152 \pm 39 \) meters for a 25km\textsuperscript{2} region. 

We observe that our architecture has memory requirements that quickly increase for higher dimensional signals. In this section, we considered 2D signals with multiple input features representing time. However. we also attempted to process this as a 3D convolution over two spatial dimensions and time. In that case, while we were able to obtain preliminary results, we encountered significant memory bottlenecks. Sparse convolutions show promise in mitigating this challenge \cite{Graham17-SubmanifoldSparse, Choy19-4DSpatioTemporal}.

%% file: journal/mid.tex
% !TeX root = journal.tex

\section{Mobile Infrastructure on Demand}\label{sec:mid}

In this section, we continue to evaluate the utility of our framework on a different problem, called mobile infrastructure on demand (MID). The overall goal in MID is to deploy a team \emph{communication agents} that provides a wireless network for a team of \emph{task agents}. Below we describe the task, summarize previous works, and describe our experiments. There, as before, Theorem \ref{thm:stationary_bound} motivates training a CNN on small examples but evaluating for large tasks.

The MID task is defined as follows. Let \( \calX = \{ \bfx \in \reals^2 \} \) be a set that represents the positions of the task agents. Denote the positions of the communication agents by a set \( \calY = \{ \bfy \in \reals^2 \} \). The positions of the task agents are exogenous. The goal is to determine communication agent positions \(\calY\) that maximize connectivity between task agents. This task is static, unlike in MTT where system dynamics are simulated.

This problem was first formulated by \cite{Mox20-MobileWirelessNetwork} as a convex optimization problem. However, the time it takes to find a solution grows quickly with the total number of agents \(|\calX|+|\calY|\). The runtime is 30 seconds for 20 agents \cite{Mox22-Learning} and can be extrapolated to 48 minutes for 100 agents and 133 \emph{hours} for 600 agents  \footnote{We obtain these estimates by fitting an exponential polynomial, and power models to data from \cite{Mox22-Learning} and reporting the lowest runtime.}. To alleviate this issue, \cite{Mox22-Learning} proposes a deep learning model that imitates the convex optimization solution. The authors utilize a convolutional encoder-decoder trained with a dataset of images representing configurations with two to six task agents uniformly distributed throughout a 320x320 meter area. 
In that scenario, the CNN achieved nearly identical performance to the convex solver while taking far less time to compute solutions. Its zero-shot performance was also evaluated on out-of-distribution configurations with up to 20 agents.
In the following set of experiments, we go one step further and test the performance on larger images that represent bigger areas and more agents. In particular, we perform MID on up to 1600x1600 meter areas and with over 600 total agents.

\subsection{Simulations}\label{subsec:mid-simulations}

We use the following method to generate task agent configurations, \(\calX\). Consider a square window with width \(T\) meters.  The number of task agents \( |\calX| \) is proportional to the window area, \(T^2\). There are five agents for \(T = 320\) and 125 agents for \(T = 1600\). We sample each task agent position independently from \( \bfx \sim U(-T/2, T/2)^2 \), a uniform distribution covering the entire window. We represent each configuration \(\calX\) by an intensity function \( X(\bft, \calX) \) following \eqref{eq:intensity_gaussian}, so that each agent is represented by a Gaussian pulse with standard deviation \( \sigma_X = 6.4 \). The image \( \bfX \in \reals^{N \times N} \) is sampled from \( X(\bft, \calX) \) at a spatial resolution of \( \rho = 1.25 \) meters per pixel.

\begin{figure*}[ht]
    \centering
    \includegraphics[width=\figWidth\linewidth]{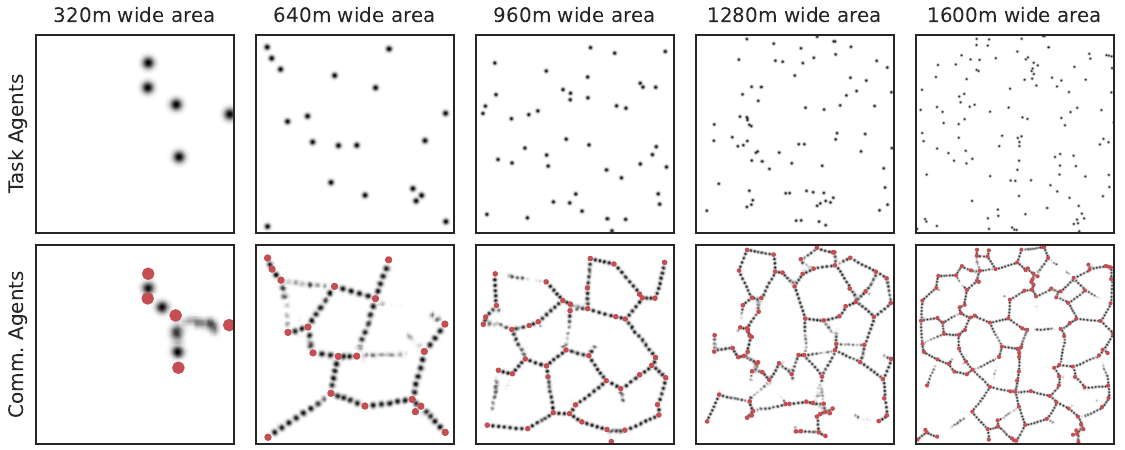}
    \caption{Example inputs and outputs to the CNN for the MID task at different window widths \( \inW = 320, 650, 960, 1280, 1600 \) but with a constant spatial resolution of \( \rho = 1.25 \) meters per pixel \cite{Owerko23-SolvingLargescale}. The top row images represent the positions of the task agents. The bottom row images represent the estimated optimal positions of the communication agents by the CNN. Additionally, the task agent positions are marked in red on the bottom images.}
    \label{fig:mid_transfer}
\end{figure*}

We use the pre-trained CNN from \cite{Mox22-Learning}. It was trained on images representing configurations within a \(T = 320\) meter wide window. It is publicly available online\footnote{\url{https://github.com/danmox/learning-connectivity}}. During inference, \( \bfX \) is the input to the CNN. The corresponding output \( \bm\Phi(\bfX; \calH) \) is assumed to represent a proposed communication agent configuration. Figure \ref{fig:mid_transfer} shows example inputs and outputs of the CNN for different window sizes.

\subsection{Baseline Approaches}

We do not compare the performance of MID to any baseline. In our experiments on the MTT task, we were able to compare the performance of the CNN against Bayesian filters. Unfortunately, this was not possible for MID due to the poor scalability of the convex optimization-based approach. Based on experimental runtimes reported in \cite{Mox22-Learning} we extrapolate that the runtime for 100 and 600 agents would be 48 minutes and 133 hours, respectivelly. Therefore, we are focused on comparing the performance of the CNN at different scales.

\subsection{Modeling Multi-Agent Communication}

To test the performance of the proposed CNN framework in MID, we measure the power required to maintain a minimum bitrate between any two agents in the network.
To do so, we assume a path loss channel model \cite{Fink11-CommunicationTeams}. It provides a relatively simple description of point-to-point communication between any two agents. The model relates three key variables: the transmit power \(P(d)\) in mili-watts, the distance \(d\) in meters, and the expected communication rate \(R\). Their relationship is summarized by \eqref{eq:minimum_power}.
\begin{equation}\label{eq:minimum_power}
    P(d) = [\text{erf}^{-1}(R)]^2 \frac{P_{N_0} d^n}{K}
\end{equation}
% {The communication rate is expressed as a fraction \( R \in [0,1] \) of some maximum channel capacity.} 
We assume that the agents' communication hardware is homogenous so that the following parameters are constant. \(P_{N_0} = 1 \times 10^{-7}\) is the noise level in the environment, \(K = 5 \times 10^{-6}\) is a parameter that describes the efficiency of the hardware, and \(n = 2.52\) characterizes the attenuation of the wireless signal with distance. Finally, we require that any two agents to communicate at a normalized rate of \(R = 0.5\). Hence, \eqref{eq:minimum_power} expresses the required transmission power as a function of distance.

Equation \eqref{eq:minimum_power} computes the power needed for agents to communicate directly. However, not all agents need to communicate directly. In many configurations, it is more efficient if messages are routed along multi-hop paths between agents. To measure the power needed to maintain this we introduce the \emph{average minimum transmit power} (AMTP).
\begin{definition} (AMTP) \label{def:amtp}
    Consider a fully connected graph between all agents. Let \( P(d) \) be the edge weights following \eqref{eq:minimum_power} with constants \(P_{N_0}\), \(K\), \(n\), and \(R\). Now, consider the minimum spanning tree of this weighted graph. We define the AMTP as the average edge weight of this tree.
\end{definition}
Notice that in Definition \ref{def:amtp} there is a path between any two agents. Each pair of agents along this path can communicate at a rate of \(R\). Therefore, the communication rate along this path is \(R\). We assume that overhead is negligible. Therefore AMTP quantifies the average transmit power needed to sustain the minimal communication rate.

We use AMTP to evaluate the performance of the proposed CNN. Recall that the output of the CNN is an image \( \bm\Phi(\bfX, \calH) \). To compute AMTP, we estimate the communication agent positions using the procedure outlined in Section \ref{sec:representing}. The result is a set of vectors \(\calY = \{ \bfy \in \reals^2 \} \), which, along with \(\calX\), is used to compute AMTP. 

The authors of \cite{Mox22-Learning} measure CNN performance via algebraic connectivity of a communication network with fixed transmission power. Algebraic connectivity is defined as the second smallest eigenvalue of the graph Laplacian.
There are two main reasons to use AMTP instead.
First, algebraic connectivity depends on the number of nodes in a graph. This complicates comparing networks of different sizes.
Second, when the transmission power is fixed, the communication graph can be disconnected. 
In those cases, the connectivity is zero, even if the graph is almost connected. This is especially problematic for large graphs because it only takes a single out-of-place agent. AMTP does not require any normalization for different graph sizes and takes into account how close a communication network is to being connected.

\subsection{Results}
To test the performance of the proposed convolutional approach in large-scale MID and the applicability of Theorem \ref{thm:stationary_bound}, we conduct the following experiments. We have a model that was trained on a \( T = 320 \) meter wide area and evaluate its performance on varying window sizes with \( T = 320, 480, 960, 1280, 1600 \) meters. At each window size, we generate 100 random task agent configurations following Subsection \ref{subsec:mid-simulations}. For each configuration, the CNN produces a communication agent configuration.

\begin{table}
    \centering
    \caption{Numerical results at different window widths with 100 samples each.}
    \label{tab:mid_power}
    \begin{tabular}{@{}rrrcc@{}}
        \toprule
        \multirow{2}{*}{Window width (m)} & \multirow{2}{*}{Task Agents} & \multirow{2}{*}{Comm. Agents} & \multicolumn{2}{c}{AMTP (mW)}        \\
        \cmidrule{4-5}
                                          &                              &                               & Mean                          & STD  \\
        \midrule
        320                               & 5                            & 12.57                         & 16.60                         & 3.21 \\
        640                               & 20                           & 68.75                         & 18.30                         & 4.54 \\
        960                               & 45                           & 165.79                        & 18.27                         & 2.71 \\
        1280                              & 80                           & 308.85                        & 18.48                         & 2.06 \\
        1600                              & 125                          & 494.24                        & 18.03                         & 1.37 \\
        \bottomrule
    \end{tabular}
\end{table}

We evaluate the AMTP for the combined communication network that includes both the task and communication agents. Table \ref{tab:mid_power} summarizes the numerical results. The AMTP increases modestly with scale. The AMTP is lowest for a 320-meter window width. It rises to the highest level when \(T = 1280\)m, rising to 11.33\% higher than at \(T = 320\)m. Interestingly, we observe a decrease at \(T = 1600\) to only 8.61\% higher than at \(T = 320\)m. Simultaneously the AMTP variance decreases as the scale increases. Combined, these two observations are strong evidence that the AMTP will plateau at this level.

\begin{figure}
    \centering
    \includegraphics[width=\figWidth\linewidth]{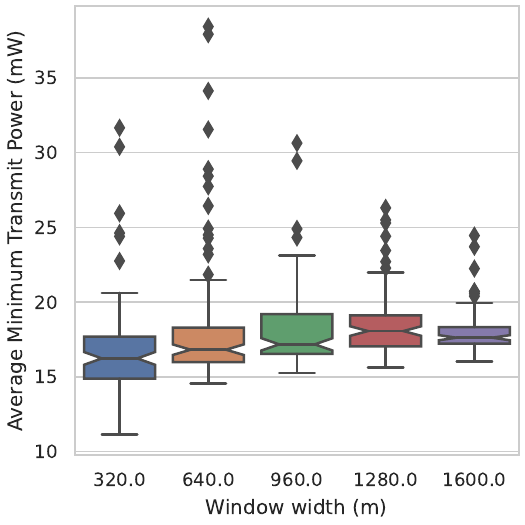}
    \caption{The distributions of the minimum transmitter power needed to maintain a normalized communication rate of at least 50\% between any two agents \cite{Owerko23-SolvingLargescale}. The distributions for each window width are visualized by box plots, with notches representing a 95\% confidence interval for the estimate of the median.}
    \label{fig:mid_power}
\end{figure}

Additionally, we think there is a quite convincing explanation for the trend in variance. The rapid decrease in variance is acutely visualized in Figure \ref{fig:mid_power}. There could be two factors working together. Recall that AMTP is effectively the average transmission power over the edges of a minimum-spanning tree. Therefore, by the Central Limit Theorem, the variance should be inversely proportional to the number of edges in the graph. However, the variance decreases faster than that. Therefore we think there is a second factor involved: border effects due to padding. These contribute to variability but would be less pronounced as the area grows relative to the perimeter.

Overall, we observe that the proposed CNN model is a great solution to large-scale MID, as it demonstrates low levels of AMTP in all settings. Unlike in Section \ref{sec:mtt} we are unable to compute the MSE, because a convex optimization solution was computationally intractable at large scales. Although, Theorem \ref{thm:stationary_bound} is a bound on the MSE, not the AMTP, good AMTP generalization is strongly aligned with the theorem as there is only a modest deterioration in performance as the window size increases.

% Nevertheless, our results only support good generalization performance when measured locally. Our theoretical result is limited to MSE performance. In our experiments, OSPA and AMTP are both averages of a local metric. Therefore, there is no evidence that tasks that require more global coordination would be similarly behaved. For instance, it is unlikely that this framework would be applicable to image classification. On the other hand, similar results in image segmentation are expected. Overall, there exists a broad class of problems where this method would be useful.

%% file: journal/conclusions.tex
\section{Conclusions}

This paper presented a transfer learning framework that allows for efficient training of CNNs for large-scale translationally symmetric problems. Training is performed on a small window of the larger problem and the trained network is transferred to much larger areas. The proposed approach is theoretically justified, by analyzing the CNN behavior with joint stationary input-output signals. The novel analysis proves that the MSE for the large-scale problem is bounded above by the MSE for the small-scale problem and a small term.  
This term results from errors introduced by zero-padding before convolutions, but it is equal to zero when padding is not used.
The novel convolutional framework was evaluated through simulations on two distinct problems: Multi-target tracking, and Mobile-infrastructure on demand. In both tasks, the proposed architecture showcased remarkable performance and was able to generalize to large-scale problems with high performance.

%% file: journal/appendix.tex
% !TeX root = journal.tex

\section{Proof of Theorem \ref{thm:stationary_bound}}\label{appendix:stationary_bound}
In this section, we derive an upper bound for \eqref{eq:mse_inf} and prove Theorem \ref{thm:stationary_bound}. For concisenes, we make the substitution \( \varepsilon(\bft) = \bm\Phi(X; \calH)(\bft) - Y(\bft) \).
\begin{align}
    \Linf(\calH) &= \E{ \avg{ |\varepsilon(\bft)|^2 } }\\
    \intertext{
    Recall that \(\calC_T^d \subset \reals^d \) is a \(T\)-wide hypercube centered at zero.\endgraf
    Next, we construct a sequence that bounds the limit above. For any \(T\) let \(N = \lceil \frac{T}{\outW} \rceil\). When we substitute for \(T\) we can make an upper bound.
    }
    \Linf(\hat\calH) & \le \E{ \lim_{N \to \infty} \frac{1}{(N\outW)^d}
        \int_{\calC_{N\outW}^d} |\varepsilon(\bft)|^2 d\bft
    } \label{eq:proof_limit_sequence}                                                                                                         \\
    \intertext{
    Notice that we can partition the hypercube \(\calC_{N\outW}^d\) into smaller hypercubes. Since \(N\) is integer, we can evenly arrange \(N^d\) hypercubes with side length \(\outW\) within it. Let \( \calC_{\outW}^d(\bm\tau) \) be such a hypercube centered at \(\bm\tau\). To partition \( \calC_{N \outW}^d \) they must be centered at}
    \calT &= \left\{i\outW - \frac{(N-1)\outW}{2} \mid i \in \bbZ, 0 \le i < N \right\}^d.\\ 
    \intertext{Thus, we write the original hypercube as 
    \( \calC_{N\outW}^d = \bigcup_{\bm\tau \in \calT} \calC_{\beta}^d(\bm\tau) \)
    and break-up the integral in \eqref{eq:proof_limit_sequence} into \(N^d\) smaller integrals.
    }
    \Linf(\hat\calH) & \le \E{ \lim_{N \to \infty} \frac{1}{(N\outW)^d} \sum_{\bm\tau \in \calT}
        \int_{\calC_{\outW}^d} |\varepsilon(\bft - \bm\tau)|^2 d\bft
    }                                                                                                         \\
    \intertext{
    Above, instead of shifting the hypercube by \(\bm\tau\) we equivalently shift the integrand.   
    By Assumption \ref{assume:processes_bounded}, the stochastic processes are bounded. 
    Therefore, both the limit and expectation exist and are finite, so we can exchange them.}
    \Linf(\hat\calH) & \le \lim_{N \to \infty} \frac{1}{(N\outW)^d} \sum_{\bm\tau \in \calT} \E{
        \int_{\calC_{\outW}^d} |\varepsilon(\bft-\bm\tau)|^2 d\bft
    }                                                                                                         \\
    \intertext{
    \endgraf
    In \cite{Zou20-LipschitzBounds}, the authors show that if the input to a CNN is strictly stationary then so is the output. Their result can be extended to jointly stationary signals as follows. Consider a vector valued process \( Z(\bft) = [X(\bft), Y(\bft)] \) that is a concatenation of \(X\) and \(Y\). Since \(X\) and \(Y\) are jointly stationary then \(Z\) is stationary. Notice that for any CNN \( \bm\Phi(X; \calH) \), we can define a new CNN with output \( \hat{\bm\Phi}(Z; \calH)(\bft) = [\bm\Phi(X; \calH)(\bft), Y(\bft)] \). According to \cite{Zou20-LipschitzBounds} this output is stationary whenever \(Z\) is stationary and the difference \( \varepsilon(\bft) = \bm\Phi(X; \calH)(\bft) - Y(\bft) \) must also be stationary. 
    \endgraf
    The expectation of a stationary signal is shift-invariant, meaning that \( \E{|\varepsilon(\bft-\bm\tau)|^2} = \E{|\varepsilon(\bft)|^2} \).
    }
    \Linf(\hat\calH) & \le \lim_{N \to \infty} \frac{1}{(N\outW)^d} \sum_{\bm\tau \in \calT} \E{
        \int_{\calC_{\outW}^d} |\varepsilon(\bft)|^2 d\bft
    }
    \intertext{The summand no longer depends on \( \bm\tau \). Since the cardinality of \(\calT\) is \( N^d \), this cancels out with the factor of \(N^d\) in the denominator. The limit expression no longer depends on \(N\), so the right hand side reduces to an average over \( \calC_\outW^d \). }
    \Linf(\hat\calH) & \le \frac{1}{\outW^d} \E{ \int_{\calC_\outW^d} |\varepsilon(\bft)|^2 d\bft } \\
    \intertext{
    \endgraf
    At this point, it is convenient to change notation. Let \( \sqcap_s(\bft) \) be an indicator function that is one whenever \( \bft \in \calC_s^d \) for all \(s \in \reals\).
    Recognize that the integral above is the L2 norm squared of \(\varepsilon(\bft)\) multiplied by a window, \(\sqcap_\outW(\bft)\).
    }
    \Linf(\hat\calH) & \le \frac{1}{\outW^d} \E{ \norm{ \sqcap_\outW (\bm\Phi(X; \hat\calH) - Y) }_2^2  }                       \label{eq:norm_without_limit}\\
    \intertext{
    The output of \(\bm\Phi(X;\hat\calH)\) is \(x_L\), the output of the \(L^\text{th}\) layer. Similarly, denote \(\tilde{x}_L\) as the output of the \(L^\text{th}\) layer in \(\bm\Phi(\sqcap_\inW X; \hat\calH)\). Hence, write the norm in \eqref{eq:norm_without_limit} as \({|| \sqcap_\outW( x_L - \tilde{x}_L + \tilde{x}_L - Y ) ||_2^2}\) and apply the triangle inequality.
    }
    \Linf(\hat\calH) & \le
            \frac{1}{\outW^d} \E{\norm{\sqcap_\outW \tilde{x}_L - Y)}_2^2}                                      \nonumber\\
        &+  \frac{1}{\outW^d} \E{\norm{ \sqcap_\outW (x_L - \tilde{x}_L) }_2^2}                              \nonumber\\
        &+ \frac{1}{\outW^d} \E{\norm{\sqcap_\outW (\tilde{x}_L - Y)}_2\norm{ \sqcap_\outW (x_L - \tilde{x}_L) }_2 } \\
    \intertext{
    Let us consider the expectation of each term individually. 
    The first term is equal to \( \Lfin(\hat\calH) \) as defined in \eqref{eq:mse_window}; this is easy to see by substituting \(\tilde{x}_L = \bm\Phi(\sqcap_\inW X; \hat\calH) \). 
    The second term is bounded above following Lemma \ref{lemma:induction}. We define \(H \defeq \prod_{l=1}^L \norm{h_l}_1\) to express this concisely.
    Finally, the third term can be bounded above by applying Cauchy-Schwartz to the expectation. That is, \( \E{XY}^2 = \E{X^2}\E{Y^2} \) where \(X,Y\) are dummy random variables.
    }
    \Linf(\hat\calH) & \le \Lfin(\hat\calH) 
        + \frac{H^2}{\outW^d} \E{\norm{ \sqcap_\outW (X - \sqcap_\inW X) }_2^2} \nonumber\\
        &+ \sqrt{ \Lfin(\hat\calH) \frac{H^2}{\outW^d} \E{\norm{ \sqcap_\outW (X - \sqcap_\inW X) }_2^2} }  \\
    \intertext{
    We can further evalaute the expectation. Since \(X\) is stationary, the second term is just the variance of \(X\) times the volume in \( \calC_{\outW+LK}^d \setminus \calC_\inW^d \). Both hypercubes are centered around zero, so the net hypervolume is the difference between their volumes. If \(\outW+LK \ge \inW\) then the hypervolume is \( (\outW+LK)^d - \inW^d \), otherwise it is zero.
    }
    \Linf(\hat\calH) & \le \Lfin(\hat\calH) + \E{X(0)^2} C + \sqrt{\Lfin(\hat\calH) \E{X(0)^2} C} \\
    C &= \frac{H^2}{\outW^d} \max\left(0, [\outW + LK]^d - \inW^d\right)
\end{align}

\section{Proof of Lemma \ref{lemma:induction}}

\begin{lemma}\label{lemma:induction}
    Let \(x_l\) denote the output of the \(l^\text{th}\) layer of \(\bm\Phi(X; \calH)\) with \(L\) layers and filter width \(K\).
    Similarly, let \(\tilde{x}_l\) denote the layers' outputs in \(\bm\Phi(\sqcap_\inW X; \calH)\). Then for all integers \(L \ge 0\) the following inequality holds.
    \[\norm{\sqcap_{\outW} (x_L - \tilde{x}_L)}_2
        \le \norm{ \sqcap_{\outW+LK} ( X- \sqcap_\inW X) }_2 \prod_{l=1}^{l=L} \norm{h_l}_1 \]
\end{lemma}

\newcommand{\LHS}{\Delta_{L+1}}
\begin{proof} We proceed by induction. For \(L=0\) the above inequality holds by definition. Hence, assume that the inequality holds for some \(L > 0\) and consider the left hand side for \(L+1\), which we denote \(\LHS\).
    \begin{align}
        \LHS & \defeq \norm{\sqcap_\outW (z^{L+1} - z^{L+1}_\inW)}_2                                                            \\
        \intertext{Then by definition of the CNN we can expand the output of the first layer.}
        \LHS & = \norm{\sqcap_\outW ( \sigma(h_{L+1} * x_L) - \sigma(h_{L+1} * \tilde{x}_L) )}_2                            \\
        \intertext{Now we can use the fact that \(\sigma\) is normalized Lipshitz from assumption \ref{assume:lipschitz}.}
        \LHS & \le \norm{\sqcap_\outW h_{L+1} * (x_L - \tilde{x}_L) }_2                                                     \\
        \intertext{Now notice that for all \(t\) such that \(|t| \le \outW/2\) the values of \((h_1*x_L)(t)\) depend only on \(x_L(s)\) for \(|s| < (\outW+K)/2\) because \(h_L\) has width \(K\). Using the monotonicity of the norm, we can simultaneously remove the \(\sqcap_\outW\) outside the convolution.}
        \LHS & \le \norm{h_{L+1} * [\sqcap_{\outW+K} (z^{L-1} - z^{L-1}_\inW)] }_2                                       \\
        \intertext{Therefore we can apply Young's convolution inequality.}
        \LHS & \le \norm{h_{L+1}}_1 \norm{\sqcap_{\outW+K} (x_L - \tilde{x}_L)}_2                                           \\
        \intertext{Now assuming the proposition holds for \(L\) then using the substitution \(\hat\outW = \outW + K\) we can obtain the desired form.}
        \LHS & \le \norm{h_{L+1}}_1 \norm{\sqcap_{\hat\outW+LK} ( X- \sqcap_\inW X) }_2 \prod_{l=1}^{l=L} \norm{h_l}_1 \\
        \LHS & \le \norm{\sqcap_{\outW+LK} ( X- \sqcap_\inW X) }_2 \prod_{l=1}^{l=L+1} \norm{h_l}_1
    \end{align}
    Thus, we have proven the inequality by induction for any integer \(L \ge 0\).
\end{proof}